\RequirePackage[T1]{fontenc}
\documentclass[journal,10pt]{IEEEtran}
\usepackage{cite}
\usepackage{amsmath,amssymb}
\usepackage{graphicx}
\usepackage{booktabs,array,tabularx}
\usepackage{url}
\usepackage{placeins}
\usepackage[hidelinks]{hyperref}

\newcommand{\method}{HTCE}

\newcommand{\ind}{\mathbb{I}}
\newtheorem{proposition}{Proposition}
\newtheorem{definition}{Definition}
\newcolumntype{Y}{>{\raggedright\arraybackslash}X}
\hypersetup{
  pdftitle={Homeomorphism-inspired Topological Cross-Scene Experience Canonicalization and Reuse for Object Navigation},
  pdfauthor={Xiaoran Xu, Yupeng Wu, Tianyu Xue, Xuanran Dong, Yifan Xu, Xiaoshan Yang, Changsheng Xu}
}
\title{HTCE: Homeomorphism-inspired Topological Cross-Scene Experience Canonicalization and Reuse for Object Navigation}

\author{Xiaoran~Xu,
        Yupeng~Wu,
        Tianyu~Xue,
        Xuanran~Dong,
        Yifan~Xu,
        Xiaoshan~Yang\textsuperscript{\ensuremath{\dagger}},
        and~Changsheng~Xu, ~\IEEEmembership{Fellow,~IEEE}

\thanks{X. Xu, Y. Wu, Y. Xu, X. Yang, and C. Xu are with the MAIS, Institute of Automation, Chinese Academy of Sciences, Beijing, China, and also with the School of Advanced Interdisciplinary Sciences, University of Chinese Academy of Sciences, Beijing, China (e-mail: xuxiaoran22@mails.ucas.ac.cn; wuyupeng23@mails.ucas.ac.cn;yang.xiaoshan3@gmail.com;\{yifan.xu, changsheng.xu\}@nlpr.ia.ac.cn).}
\thanks{T. Xue and X. Dong are undergraduates at Tsinghua University and University of Chinese Academy of Sciences, respectively.}%
\thanks{\textsuperscript{\ensuremath{\dagger}}Corresponding author: Xiaoshan Yang.}%
}

\begin{document}
\maketitle
\begin{abstract}
As embodied agents perform object-goal navigation across residential environments, they accumulate successful experiences whose semantic structure recurs across scenes even when geometric trajectories do not. Reusing such experience is challenging because differences in region naming and observation granularity produce distinct graph representations of similar approaches, while meaningful differences in anchors, branches, and endpoints must stay distinguishable. We introduce \method{}, a homeomorphism-inspired topological framework that canonicalizes and reuses navigation experience. Feature and topology subdivisions characterize representational variations, and their semantically constrained reductions produce comparable semantic sequences and structural signatures while preserving semantic anchors, branching relations, and navigation endpoints. \method{} then retrieves historical successful approaches through subdivision-aware matching as guidance for a frozen vision-language policy. Canonicalization compresses the experience index losslessly, merging 17.7--23.6\% of distinct stored route descriptions while retaining full task performance. Under a single frozen vision-language backbone shared by all compared systems, \method{} attains the best SR and SPL in every setting---59.9\%/31.8\% on HM3D v0.1, 71.0\%/31.5\% on HM3D v0.2, and 46.6\%/16.6\% on MP3D---with SR gains of up to 4.9 points and structured retrieval contributing up to 2.8 SPL points. The code is available in https://github.com/startracker0/htce.git
\end{abstract}

\section{Introduction}
\label{sec:introduction}

\IEEEPARstart{O}{bject}-goal navigation requires embodied agents to locate target objects in unfamiliar environments using partial observations \cite{anderson2018evaluation,batra2020objectnav,majumdar2022zson,nie2025wmnav,chaplot2020object,gervet2023navigating}. As agents navigate across different residential environments, they can accumulate successful experiences that record target encounters and the semantic approaches leading to them \cite{khanna2024goatbench,wang2026trajrag}. Although these environments differ in appearance and geometric layout, successful navigation approaches may exhibit recurring semantic structures, such as reaching a target in a bedroom through a passage from a living room \cite{wu2019bayesian}. Such similarities suggest that historical navigation experiences may provide useful guidance for object searches in new environments.

Establishing such correspondences remains challenging because semantically similar navigation approaches may yield different graph representations. A passage described as a hallway in one scene may appear as an entryway followed by a corridor in another, while repeated observations may introduce additional intermediate vertices. These variations affect semantic descriptions and topological granularity without necessarily altering the underlying navigation structure. Conversely, changes in meaningful semantic regions, branching connections, or goal endpoints may distinguish different navigation approaches. Therefore, cross-scene experience matching should accommodate representational variations while preserving navigation-relevant structural distinctions.

\begin{figure*}[t]
\centering
\includegraphics[width=\textwidth]{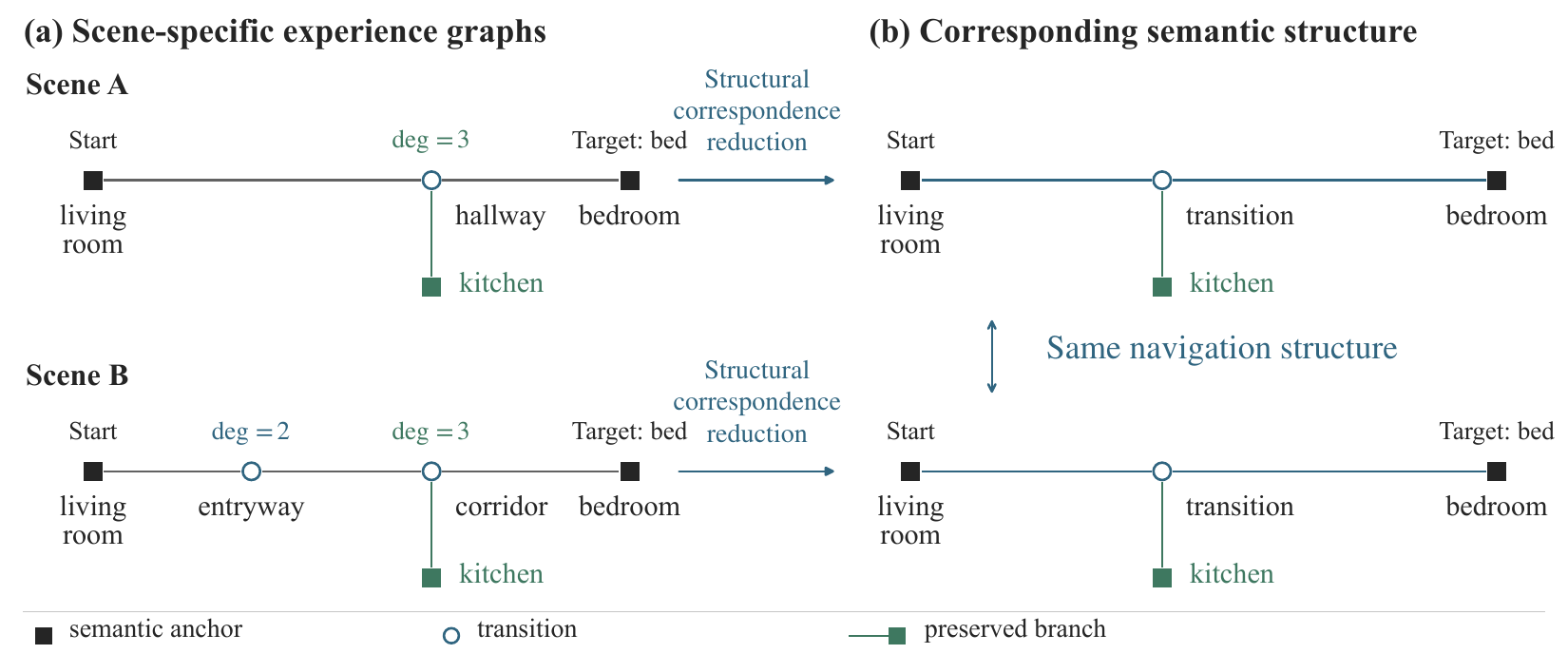}
\caption{Different scene representations, corresponding navigation structure. (a) Two schematic experience graphs describe a bed search from a living room to a bedroom. Scene B introduces a degree-two entryway before the corridor junction. (b) Structural correspondence reduction aligns compatible passage labels and suppresses the entryway, preserving the start and target endpoints, semantic anchors, and kitchen branch. Both reduced graphs describe a living-room-to-bedroom approach with a kitchen branch at the intervening junction. Suppressed-transition counts remain available to graded matching: the living-room-to-junction connections contain zero and one suppressed vertices, respectively.}
\label{fig:motivation}
\end{figure*}

Graph homeomorphism offers a natural perspective for handling topological variations, as edge subdivision introduces intermediate degree-two vertices without changing the underlying connectivity \cite{diestel2017}. Applying this principle to navigation experience graphs requires additional semantic constraints, since not all intermediate vertices can be treated as removable subdivisions. For example, transition vertices along nonbranching passages may be suppressed, whereas semantic regions, junctions, and navigation endpoints must remain explicit. Building on this principle, we introduce \method{}, a homeomorphism-inspired topological framework for cross-scene navigation experience canonicalization and reuse.
The framework characterizes representational variations through two complementary operations: feature subdivision, which captures compatible semantic descriptions, and topology subdivision, which accounts for intermediate transition vertices along nonbranching passages. Through feature canonicalization and semantically constrained topological reduction, \method{} constructs comparable semantic sequences and structural signatures while preserving semantic anchors, branching relations, and navigation endpoints. As illustrated in Fig.~\ref{fig:motivation}, approaches represented at different granularities can be structurally aligned without removing meaningful branches.

We further integrate \method{} with persistent navigation memory to support cross-scene experience reuse \cite{zheng2024esceme,wang2026trajrag}. Across successive navigation episodes, the agent stores successful navigation approaches and associated target observations. Based on the reduced representations, subdivision-aware approximate matching retrieves relevant historical experiences as semantic guidance for a frozen vision-language model (VLM) \cite{bai2025qwen3vl}, while action selection remains grounded in current visual observations \cite{nie2025wmnav,goetting2025end}. We report navigation comparisons on HM3D and MP3D \cite{ramakrishnan2021hm3d,yadav2023hm3dsem,chang2017matterport3d,savva2019habitat} and conduct recorded-route and controlled-graph analyses to examine the structural properties of the proposed representation.

A canonicalization that preserves navigation structure should deliver compression without loss, and we establish both halves. Collapsing admissible subdivisions makes stored experiences comparable and the index compact, while the information the policy acts on is retained intact: reduction merges distinct descriptions of the same approach and leaves task performance at full strength. This is lossless compression of navigation experience, and it is what makes a single canonical index reusable across scenes.

Our main contributions are summarized as follows:
\begin{itemize}
\item \textbf{Homeomorphism-inspired canonicalization.} We introduce a homeomorphism-inspired topological framework that characterizes compatible feature and topology subdivisions and defines structural correspondence between navigation experiences while preserving semantic anchors, branching relations, and navigation endpoints.
\item \textbf{Subdivision-aware graph matching.} We develop feature canonicalization and semantically constrained topological reduction to construct comparable semantic sequences and structural signatures for approximate matching of navigation approaches.
\item \textbf{Cross-scene navigation experience reuse.} We integrate subdivision-aware matching with persistent navigation memory and a frozen vision-language model, and show that canonicalization compresses the experience index losslessly: it merges 17.7--23.6\% of stored descriptions and satisfies every prescribed invariance and sensitivity relation while retaining full task performance, with structured retrieval contributing up to 2.8 SPL points.
\end{itemize}

\section{Related Work}
\label{sec:related}

\subsection{Object Navigation With Foundation Models}
Foundation models support object navigation by connecting visual observations with semantic goals \cite{zhou2023esc,huang2023vlmaps}. L3MVN and OpenFMNav leverage language and vision-language priors for exploration \cite{yu2023l3mvn,kuang2024openfmnav}, while VLFM uses visual-semantic relevance to guide frontier exploration \cite{yokoyama2024vlfm}. VLMNav and WMNav further employ vision-language reasoning for navigation decisions \cite{goetting2025end,nie2025wmnav}. Rather than introducing a new navigation policy, \method{} provides a frozen VLM with structural guidance retrieved from successful historical experiences.

\subsection{Navigation Memory, Scene Graphs, and Experience Retrieval}
\label{sec:related_memory}
Topological memories represent spatial connectivity for navigation \cite{savinov2018semiparametric,an2025etpnav,chaplot2020neural}. These methods maintain graph structures over observations but do not explicitly handle variations in region naming or observation granularity across scenes. Semantic scene graphs further support object- and region-level reasoning \cite{yin2024sgnav,loo2025openscenegraphs,armeni2019scene,gu2024conceptgraphs}. UniGoal employs scene-goal graph matching for exploration, and SSMG-Nav organizes persistent semantic memory through spatial keypoints \cite{yin2025unigoal,niu2026ssmgnav}. Navigation memory extends perception beyond current observations \cite{zheng2024esceme,li2024memonav,zhang2025mem2ego,fang2019scene}. GOAT-Bench evaluates memory use across successive navigation goals \cite{khanna2024goatbench}, while TrajRAG retrieves historical geometric-semantic trajectories through hierarchical retrieval with a trained trajectory encoder \cite{wang2026trajrag}. Beyond navigation, graph edit distance supports comparison between nonidentical graphs \cite{gao2010survey,riesen2009approximate}, and graph homeomorphism characterizes connectivity preserved under edge subdivision \cite{diestel2017}.

Several recent methods accumulate experience across episodes or scenes. GOAT-Bench benchmarks lifelong multi-goal navigation with persistent memory \cite{khanna2024goatbench}. TrajRAG retrieves cross-episode geometric-semantic trajectories using contrastive-trained embeddings \cite{wang2026trajrag}. C-NAV addresses continual object navigation in open worlds with self-evolving category knowledge \cite{yu2025cnav}, and EvolveNav uses proactive preflection and self-evolving memory for zero-shot navigation \cite{chai2026evolvenav}. Unlike these approaches, \method{} uses homeomorphism-inspired topological canonicalization to align different-granularity graph representations before matching: compatible passage labels are normalized and intermediate degree-two transitions are suppressed, producing a canonical form that absorbs naming and observation-frequency variations while preserving meaningful semantic anchors, branching relations, and navigation endpoints. This makes cross-scene matching invariant to representational differences that do not alter the navigation structure. The matcher is also comparatively light: UniGoal drives its scene--goal graph matching with a LLaVA, LLaMA, and CLIP stack, and TrajRAG trains a contrastive trajectory encoder on top of GroundingDINO, MobileSAM, and a Qwen3-32B planner, whereas \method{} performs training-free matching over a single frozen VLM backbone shared by all prompt roles.

\begin{figure*}[t]
\centering
\includegraphics[width=\textwidth]{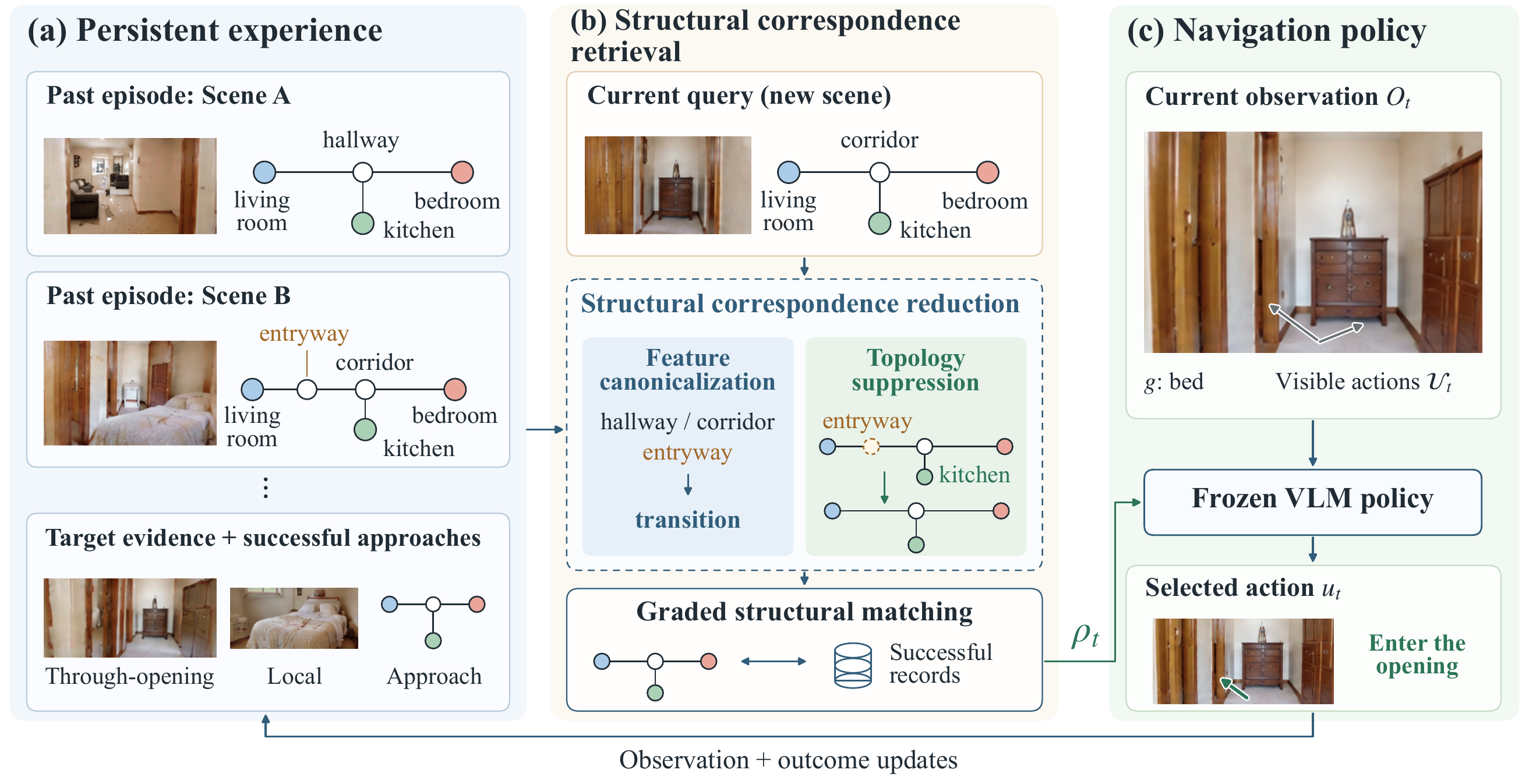}
\caption{\method{} converts persistent navigation experience into guidance for the current scene. (a) Historical records retain semantic connectivity, successful approaches, and local or through-opening target evidence. (b) Structural correspondence reduction aligns compatible passage descriptions and suppresses the degree-two entryway while preserving the kitchen junction and semantic endpoints. Graded structural matching with target evidence produces guidance $\rho_t$. (c) Current observation $O_t$, goal $g$, visible candidates $\mathcal U_t$, and retrieved guidance enter the frozen VLM policy, which selects action $u_t$. Subsequent observations and terminal outcomes update memory. Scene images, graph examples, and action overlays are illustrative.}
\label{fig:method}
\end{figure*}

\section{Preliminary}
\label{sec:setting}

\subsection{VLM Backbone}
Each object-navigation episode specifies an indoor scene, an initial state, and a target category $g$. At time $t$, the agent receives an RGB-D observation $O_t$ and local pose information. A depth-based action proposer generates visible action candidates $\mathcal U_t$, and the controller executes the candidate selected by the policy \cite{goetting2025end}. A single frozen VLM backbone handles perception, prediction, planning, and action selection through separate prompt contexts \cite{nie2025wmnav}. The backbone parameters $\theta$ remain fixed throughout all episodes; this work does not fine-tune the VLM.

Let $z_t$ denote the extracted visual-semantic record. Memory updating, experience retrieval, and action selection are expressed as
\begin{align}
\mathcal M_t^+ &= \Phi_{\mathrm{obs}}(\mathcal M_t,z_t), \label{eq:memory_update}\\
\rho_t &= \mathcal R(\mathcal M_t^+,g,z_t), \label{eq:retrieval_interface}\\
u_t &= \mu_\theta(O_t,\mathcal U_t,g,\rho_t),\qquad \theta\ \text{fixed}.
\label{eq:policy}
\end{align}
Here $\Phi_{\mathrm{obs}}$ writes the record into memory $\mathcal M_t$, $\mathcal R$ retrieves guidance $\rho_t$ containing target evidence and semantic navigation guidance, and $\mu_\theta$ is the frozen VLM policy with parameters $\theta$. Executable actions remain grounded in current observations.

\subsection{Cross-Scene Experience Reuse Protocol}
\label{sec:reuse_protocol}
Navigation episodes are processed across different scenes with persistent experience memory \cite{khanna2024goatbench,wang2026trajrag,yu2025cnav}. Episode reset reinitializes the active region chain, previous action, and odometry. Persistent snapshots store region observations, graph connections, target evidence, and successful approaches; their visibility in the next episode follows the configured load and save lifecycle. Successful approaches are associated with target records using terminal evaluator feedback.

Multiple workers process assigned episode sequences concurrently. Each maintains a working graph and exchanges persistent snapshots at episode boundaries. Appendix~\ref{app:evaluation} specifies episode scheduling and the snapshot protocol.

During navigation, the current region assignment establishes a target-conditioned approach in the stored semantic graph. Graded retrieval compares this approach with historical successful records and their descriptors. Retrieved experiences provide semantic regions, passage relations, and target-observation evidence, while current RGB-D observations determine executable actions. The policy-context interface supplies the selected cue to the VLM, and evidence gates control its use in action selection.

\subsection{What an Experience Record Contains}
The persistent memory is represented as
\begin{align}
\mathcal M &= \bigl(\mathcal G,\;\{\mathcal O_v^{\mathrm{in}},\mathcal O_v^{\mathrm{rem}}\}_{v\in V},\;\Omega\bigr), \label{eq:memory}\\
\mathcal G &= (V,\,E). \notag
\end{align}
Here $\mathcal G$ is an undirected semantic region graph whose vertices and edges are defined in Section~\ref{sec:overview}. $\mathcal O_v^{\mathrm{in}}$ and $\mathcal O_v^{\mathrm{rem}}$ store local and through-opening target observations, respectively. $\Omega$ records episode outcomes, including successful approach routes $\pi$ and their structural signatures.

Object records retain confidence, visual summaries, and viewpoint metadata, with through-opening observations additionally recording their viewing origins and inferred target-region links. Successful target records store ordered approaches, canonical role sequences, and available structural signatures. These records jointly support target-conditioned retrieval and subdivision-aware graph matching.

\section{Homeomorphism-inspired Representation}
\label{sec:method}

This section builds the canonical representation in three steps that mirror the homeomorphism analogy. \emph{Feature subdivision} and \emph{topology subdivision} (Sections~\ref{sec:route}--\ref{sec:suppression}) model the two ways scenes describe one approach differently: compatible wording of a region, and intermediate observations along a passage. Each has an inverse \emph{reduction} that removes the variation while preserving semantic anchors, branches, and endpoints. \emph{Structural correspondence} (Section~\ref{sec:correspondence}) then defines when two reduced experiences count as the same. Reduction yields two signatures used for matching (Section~\ref{sec:retrieval}): a canonical feature word $s(\pi)$ comparing routes as role sequences, and a reduced-graph signature $\Gamma$ comparing them as structures.

\subsection{Navigation Experience Graphs}
\label{sec:overview}
We represent navigation experience by an attributed graph $\mathcal E=(\mathcal G,f,\pi,\mathcal O)$, where $\mathcal G=(V,E)$ is an undirected region graph. Each vertex $v\in V$ corresponds to a region observation, and an edge $(v_i,v_j)\in E$ is created when two regions are observed in succession or through a spatial connection. The function $f$ assigns an observed region feature to each vertex, $\pi=(v_0,\ldots,v_L)$ records an ordered approach with start endpoint $v_s=v_0$ and target endpoint $v_g=v_L$, and $\mathcal O$ contains the associated target observations. Object records retain their confidence, viewing origin, visual summary, and successful approaches.

The graph distinguishes vertex identity from semantic role. Different region hypotheses can share a role, and consecutive observations can repeat the same semantic description. For cross-scene comparison, descriptors express semantic roles and their connections across different vertex identities. Feature reduction compares the descriptions, while topology reduction examines the connections. Both operations preserve access to the original observation records.

\subsection{Feature Subdivision and Canonicalization}
\label{sec:route}
The framework partitions region labels into two disjoint classes: \emph{semantic anchors} that carry navigational meaning (e.g., bedroom, kitchen) and \emph{transitions} whose specific identity does not affect the route structure. The partition is configurable; in indoor navigation we instantiate the transition class as follows. Let $\nu$ normalize an observed region name, including its spelling and predefined aliases. We define the transition vocabulary
\begin{equation}
\mathcal T=\{\text{corridor},\text{hallway},\text{entryway},\text{unknown room}\}
\label{eq:transitions}
\end{equation}
and the semantic role map
\begin{equation}
r(x)=
\begin{cases}
\tau,&\nu(x)\in\mathcal T,\\
\nu(x),&\text{otherwise},
\end{cases}
\label{eq:role}
\end{equation}
where $\tau$ is the transition role and the remaining roles are semantic anchors. For a role $a$, the compatible feature class is $\mathcal F_a=\{x:r(x)=a\}$.

We define \emph{feature subdivision} as expressing one role occurrence through one or more compatible descriptions:
\begin{equation}
a\rightsquigarrow(x_1,\ldots,x_m),\qquad
x_j\in\mathcal F_a,\quad m\geq1.
\label{eq:feature_subdivision}
\end{equation}
A transition can thus be described by a hallway observation or by an entryway followed by a corridor. An anchor can similarly appear in several consecutive observations. This operation concerns the ordered feature description. The connectivity of the corresponding graph vertices is handled by topology subdivision.

Canonicalization reverses feature subdivision. We map observed features to their roles and contract maximal runs of equal roles:
\begin{equation}
\tau\tau\rightarrow\tau,\qquad aa\rightarrow a
\quad(a\text{ a semantic anchor}).
\label{eq:rewrite}
\end{equation}
The resulting word $s(\pi)$ provides the canonical feature description used in matching. A route from living room to bedroom through a hallway has the same feature word as one through an entryway followed by a corridor: living room--$\tau$--bedroom. A direct connection between the two regions remains distinct, and $a\tau a$ preserves leaving and returning to the same semantic category.

\begin{proposition}[Canonical feature word]
\label{prop:route}
For a fixed semantic role map, contracting consecutive equal roles produces a unique reduced word. Compatible label realizations and repeated observations within a role run preserve this word.
\end{proposition}
\noindent The reduction has a unique normal form because the contractions terminate and are locally confluent; Appendix~\ref{app:proof} gives the proof and the boundary cases.

\subsection{Topology Subdivision and Semantic Constraints}
\label{sec:suppression}
Topology subdivision refines an edge into a chain:
\begin{equation}
(v,v')\rightsquigarrow(v,\xi_1,\ldots,\xi_k,v'),\qquad k\geq1,
\label{eq:subdivision}
\end{equation}
where each inserted vertex $\xi_j$ has transition role, degree two, and no additional incident branch. Existing semantic labels, surrounding connections, and query endpoints are preserved. This is classical edge subdivision restricted to semantically compatible transition vertices \cite{diestel2017}.
For semantic correspondence, the canonical role word along the refined connection must also be preserved. Extending an existing transition run satisfies this condition; inserting the first passage between two directly adjacent anchors changes the semantic connector.

Its inverse suppresses intermediate transition vertices satisfying
\begin{equation}
\begin{aligned}
C(v)={}&\ind[r(f(v))=\tau]\,\ind[\deg_{\mathcal G}(v)=2]\\
&{}\cdot\ind[v\notin\{v_s,v_g\}].
\end{aligned}
\label{eq:suppress}
\end{equation}
where $\ind[\cdot]$ is the indicator function and $\deg_{\mathcal G}(v)$ is the degree of $v$ in the stored graph. The predicate $C(v)$ equals one exactly when $v$ is suppressible, that is, a non-endpoint transition lying on a nonbranching passage; only such vertices are removed. A transition connected to an off-route kitchen keeps degree above two and is therefore retained as a junction, and a bedroom is retained regardless of degree because it is a semantic anchor. Suppression extracts the structural description used for matching while retaining the original graph and observation evidence.

Let $(y_0,\ldots,y_K)$ be the route vertices that survive suppression, in route order. The descriptor of the $i$-th retained route edge, joining $y_i$ to $y_{i+1}$, records its two endpoint roles and the number $c_i$ of transitions suppressed along it:
\begin{equation}
\epsilon_i=\bigl(r(f(y_i)),r(f(y_{i+1})),c_i\bigr),
\label{eq:edge_descriptor}
\end{equation}
so $c_i$ retains observation granularity as an attribute without reinserting the suppressed vertices.
Off-route neighbors are followed through bounded chains of degree-two transitions to describe branches. The tracing budget is specified in Appendix~\ref{app:signature_config}. The resulting signature is
\begin{equation}
\Gamma(\pi,\mathcal G)=(R,N,E_\pi,B,N_s,N_g),
\label{eq:signature}
\end{equation}
where $R$ is the retained role sequence, $N$ contains node descriptors, $E_\pi$ contains route-edge descriptors, $B$ contains branch descriptors, and $N_s,N_g$ describe endpoint-neighbor roles. Retained connectivity captures the structural relation under subdivision; transition counts and semantic descriptors supply attributes for graded comparison.

\subsection{Structural Correspondence and Preserved Structure}
\label{sec:correspondence}
We compare a rooted navigation context $\mathcal Q=(H,f,\pi,g)$: an observed connected graph $H$, an ordered simple approach $\pi$, and its object goal $g$. This domain covers breadth-first search (BFS) query paths; saved trajectories with repeated visits remain available to the graded matcher. The context contains the route and its branch connections. It can be a complete experience component or a specified subgraph; cut boundaries are marked with their stored degrees and protected. Degrees are inherited from the stored graph, so cropping a route preserves observed junctions.

With $\mathcal B_H$ denoting the marked cut-boundary vertices, the protected set is
\begin{equation}
\begin{aligned}
P={}&\{v\in V(H):r(f(v))\ne\tau\ \text{or}\ \deg_{\mathcal G}(v)\ne2\}\\
&\cup\{v_s,v_g\}\cup\mathcal B_H.
\end{aligned}
\label{eq:protected}
\end{equation}
Partition the remaining vertices into maximal transition chains between protected vertices. Replacing each chain by an edge gives a reduced multigraph $\mathcal K(\mathcal Q)$ with loops and parallel connections retained. Each edge carries the canonical role word of its original chain, including its endpoints, and $\bar\pi$ marks the ordered route through the reduced graph. Thus direct anchor adjacency and an intervening passage remain distinct. The exact transition count $c_i$ is a retrieval attribute outside this structural identity.

\begin{definition}[Structural correspondence]
\label{def:sgh}
Two contexts satisfy $\mathcal Q\sim_{\mathrm{SC}}\mathcal Q'$ when $g=g'$ and there is a multigraph isomorphism
\begin{equation}
\varphi:\mathcal K(\mathcal Q)\longrightarrow \mathcal K(\mathcal Q')
\label{eq:sgh}
\end{equation}
that preserves vertex roles, boundary marks, chain words in corresponding orientations, and the ordered route: $\varphi(v_s)=v_s'$, $\varphi(v_g)=v_g'$, and $\varphi(\bar\pi)=\bar\pi'$. The mapping preserves every edge incidence, including off-route branch attachments and connection multiplicity.
\end{definition}

The isomorphism operates on the reduced multigraphs $\mathcal K$ and corresponds to homeomorphism-compatible structural agreement on the original graphs: two navigation contexts are structurally corresponding when their reduced graphs are isomorphic under the preserved attributes, which is equivalent to the original graphs being related by admissible feature and topology subdivisions.

\begin{proposition}[Preserved navigation structure]
\label{prop:topology}
For a fixed context boundary, compatible relabeling and degree-two transition subdivision that preserves each chain word leave $\mathcal K(\mathcal Q)$ unchanged up to the isomorphism in Definition~\ref{def:sgh}. Reduction is independent of suppression order and preserves reachability between protected vertices, branch incidence, and ordered route traversal.
\end{proposition}
\noindent The proof of Proposition~\ref{prop:topology} uses the unique decomposition into maximal transition chains, as detailed in Appendix~\ref{app:topology_proof}, including cyclic configurations. Appendix~\ref{app:protected_cases} explains why distinct same-role anchors and protected boundary vertices remain explicit.

\section{Experience Retrieval and Navigation}
\label{sec:retrieval}

\subsection{From Reduced Representations to Retrieval}
\label{sec:online_retrieval}
Online retrieval applies the same reductions and reads the reduced context through the descriptor $\Gamma$. For contexts related by admissible route subdivision, Proposition~\ref{prop:topology} fixes the retained vertices with their roles, degrees, and endpoint neighborhoods, so the retained role sequence, node descriptors, and endpoint-neighbor roles agree; only the transition counts $c_i$ in $E_\pi$ change. Graded matching therefore ranks structurally corresponding records by observation granularity. Degree buckets and branch sets summarize incidence more coarsely than $\mathcal K(\mathcal Q)$: branch attachment order and connection multiplicity are exact structural distinctions that $\Gamma$ can merge. Section~\ref{sec:graph_validation} measures both relations.

\subsection{Matching Through Reduced Representations}
\label{sec:matching}
The canonical feature word and reduced graph signature are extracted from the same attributed graph and share the semantic role map. Feature agreement compares $s(\pi)$ with a successful approach $s(\pi^+)$ using normalized edit similarity \cite{wagner1974string}:
\begin{equation}
S_{\mathrm{feat}}=1-\frac{D_{\mathrm{edit}}(s(\pi),s(\pi^+))}
{\max(|s(\pi)|,|s(\pi^+)|,1)}.
\label{eq:route_score}
\end{equation}
Here $D_{\mathrm{edit}}$ is the edit distance and $|\cdot|$ denotes word length. Graph agreement is a weighted combination of component similarities:
\begin{equation}
S_{\mathrm{graph}}=\sum_{i} w_i\,\phi_i,
\label{eq:graph_score}
\end{equation}
where the components $\phi_i$ compare retained role sequences (edit similarity), route edges, branches, and endpoint neighborhoods (Jaccard overlap), respectively. The fixed nonnegative weights $w_i$ sum to one; their decomposition and values are specified in Appendix~\ref{app:retrieval_config}. Two empty sets have overlap one. The comparison uses graph signatures when enabled and available, with feature-word matching as the fallback. For each candidate, the selected structural agreement is the best match among its successful records.

This matching procedure combines semantic compatibility with the structure exposed by constrained suppression. Its continuous score measures agreement of the retained descriptors. In particular, two subdivision-related approaches can share retained connectivity while receiving different scores because their edge attributes record different numbers of suppressed transitions.

\subsection{Application to Cross-Scene Object Navigation}
\label{sec:gating}
Target observations define candidate destinations. A local observation uses its region as the endpoint; a through-opening observation uses its linked target region when available and its observer region otherwise. BFS constructs a candidate approach from the current region, and the matcher compares it with successful approaches attached to the target record. For a candidate region $v$ and goal category $g$, the base evidence score $b(v,g)$ combines target confidence, observation support, successful records, and region confirmation. Local candidates take priority over through-opening candidates.

Structural agreement contributes through an evidence-gated support term:
\begin{equation}
h(v,g)=G(v,g)\,\psi(v,g).
\label{eq:structural_term}
\end{equation}
Gate $G$ admits candidates with successful support and reliable association and path evidence within the retrieval horizon. For a candidate in another region, $\psi$ combines mean path confidence $\bar q$, structural agreement $S$ ($S_{\mathrm{graph}}$, or $S_{\mathrm{feat}}$ as the fallback), reduced-route length $\ell=|s(\pi)|$, and target-role compatibility. Length attenuation favors concise supported approaches; a same-region candidate uses a dedicated support rule. Candidates are ranked by
\begin{equation}
F(v,g)=b(v,g)+\lambda h(v,g),
\label{eq:ranking}
\end{equation}
where $\lambda$ controls the contribution of structural support to target evidence. Appendix~\ref{app:retrieval_config} gives the complete scoring function and fixed configuration.

The selected observation and matched approach form a semantic cue for the frozen VLM, which expresses region roles and connections while current observations and visible candidates determine how the agent advances. Subsequent observations update the experience graph, and terminal success attaches the completed approach to its target record, connecting canonicalization to cross-scene navigation (Fig.~\ref{fig:method}).

\begin{table*}[t]
\caption{ObjectNav performance across sequential episodes (SR/SPL, \%) with a 40-step budget. All methods are implemented and evaluated by us using the same frozen VLM backbone and identical experimental settings. $*$ denotes reproduced and adapted external baselines, and TM denotes textual memory; bold indicates the best result in each column.}
\label{tab:main_results}
\centering
\small
\begin{tabular*}{\textwidth}{@{\extracolsep{\fill}}lrrrrrr@{}}
\toprule
&\multicolumn{2}{c}{HM3D v0.1}&\multicolumn{2}{c}{HM3D v0.2}&\multicolumn{2}{c}{MP3D}\\
Method & SR & SPL & SR & SPL & SR & SPL\\
\midrule
OpenFMNav$^{*}$ &52.1&23.5&37.6&14.1&27.1&7.5\\
L3MVN$^{*}$ &37.8&19.4&29.6&13.3&22.4&7.4\\
VLMNav$^{*}$ &40.1&14.9&46.5&14.0&35.2&11.0\\
WMNav$^{*}$ &55.0&31.7&70.0&30.0&43.5&15.6\\
WMNav + TM$^{*}$ &56.3&30.3&67.2&30.7&43.3&15.7\\
\method{} &\textbf{59.9}&\textbf{31.8}&\textbf{71.0}&\textbf{31.5}&\textbf{46.6}&\textbf{16.6}\\
\bottomrule
\end{tabular*}
\end{table*}

\section{Experiments}
\label{sec:experiments}
\subsection{Evaluation Setup}
\label{sec:setup}
We evaluate navigation performance on Habitat ObjectNav using HM3D v0.1, HM3D v0.2, and MP3D \cite{savva2019habitat,ramakrishnan2021hm3d,chang2017matterport3d}, with 2,000, 1,000, and 2,195 episodes, respectively. Success rate (SR) measures target completion; success weighted by path length (SPL) also accounts for path efficiency \cite{anderson2018evaluation,batra2020objectnav}. Both are percentages, and differences are percentage points. Component comparisons examine the memory design, while recorded-route and controlled-graph analyses characterize the representations used for retrieval.

\paragraph{Controlled comparison}
Every compared system, including the external baselines, is implemented and run by us on the same frozen Qwen3-VL-Plus backbone, the same depth-based action proposer, the same controller, and the same 40-step budget. Differences between rows therefore reflect memory and matching design rather than backbone capacity, prompt engineering, or interaction budget, which is the condition required to attribute a margin to experience reuse.

\paragraph{Measurement axes}
SR and SPL resolve end-to-end system comparison, where 95\% Wilson intervals span $\pm2.1$ to $\pm2.8$ points at these episode counts and the margins we report are several times that width. Properties of the representation itself, index compactness and structural fidelity, are measured directly on fixed recorded inputs and enumerated graph transformations, where the quantities are exact and independent of episode sampling. Each claim is reported on the axis that resolves it.

The navigation experiments use Qwen3-VL-Plus \cite{bai2025qwen3vl}, depth-based visible action candidates, a shared controller, and a 40-step interaction budget. Perception and action roles share the frozen backbone. Prompt context and direct action bias are enabled, with retrieval evidence and confidence gates determining when guidance is applied. Asterisks identify adapted external systems. Appendix~\ref{app:evaluation} details episode scheduling and cost measurement.

\subsection{Navigation Performance Across Sequential Episodes}
\label{sec:navigation}
In Table~\ref{tab:main_results}, \method{} attains the highest SR and the highest SPL in every one of the three settings, reaching 59.9/31.8 on HM3D v0.1, 71.0/31.5 on HM3D v0.2, and 46.6/16.6 on MP3D. Against the strongest baseline the SR gains are 4.9 points on HM3D v0.1 and 3.1 points on MP3D, both several times the interval width of their splits, and both obtained without changing the backbone, the action space, or the step budget. The advantage is broadest on the two 2,000-episode-scale splits, which carry the finest resolution and the widest scene variety, and it holds simultaneously on SR and SPL: \method{} reaches targets more often and along shorter paths, improving SPL by 1.5 points on HM3D v0.2 and 1.0 point on MP3D. Gains on both metrics at once indicate that retrieved experience shortens the search rather than trading efficiency for coverage, which is the behavior experience reuse is meant to produce.

Compared with WMNav augmented by textual memory, \method{} has SR/SPL margins of 3.6/1.5 points on HM3D v0.1 and 3.8/0.8 points on HM3D v0.2. Structurally indexed memory represents a target encounter together with its approach and viewing origin. The following comparisons examine this memory design, then test how its representations respond to changes in descriptions and graph granularity.

\subsection{Component Comparisons}
\label{sec:components}
Table~\ref{tab:ablations} switches off one component at a time. Disabling Matching, which removes successful-approach similarity and its success-record gate while retaining target evidence, association checks, and path-length support, drops SPL to 29.0/30.2 on HM3D v0.1/v0.2 against 31.8/31.5 for the full system: margins of 2.8 and 1.3 points, the largest among the structural variants.

\noindent\textit{Canonicalization compresses the index losslessly.}
Admissible subdivisions are defined so as to leave navigation-relevant structure unchanged. Suppressing them therefore changes how many stored descriptions an approach is spread across, while the approach the policy is told about stays the same. Both halves hold, and together they are the central property of the reduction.

Compression is substantial and exact. On 927 recorded successful routes from two episode orders, canonicalization merges 130 and 140 distinct stored descriptions into 107 entries each, a 17.7--23.6\% reduction of the index (Appendix~\ref{app:recorded_routes}), collapsing compatible passage labels and repeated observations while keeping destination, order, and connector distinctions explicit. A recurring approach that raw descriptions scatter across several keys becomes one entry, and every underlying evidence record is retained, which is what lets a single index serve scenes that describe the same route differently.

Task behavior is retained in full. Disabling Reduction in Table~\ref{tab:ablations} keeps lexical labels and repeated vertices and turns off transition suppression, while graph descriptors still use the semantic role map; it reaches 59.0/30.7 and 70.8/31.3 against 59.9/31.8 and 71.0/31.5 for \method{}, so \method{} is at least as strong on every metric and split while indexing a fifth fewer descriptions. Compression at full task strength is what separates a structure-preserving reduction from lossy summarization, and it is the guarantee Definition~\ref{def:sgh} and Proposition~\ref{prop:topology} are built to deliver.

Table~\ref{tab:ablations} shows what the retrieved structure is worth. Removing successful-approach matching costs 2.8 and 1.3 SPL points, the largest margin among all structural variants: structured experience drives navigation efficiency, and canonicalization is what makes that experience indexable across scenes.

\begin{table}[t]
\caption{Component ablation on HM3D (SR/SPL, \%). \textbf{G}: experience indexed in the semantic region graph rather than text notes. \textbf{P}: local and through-opening sightings kept separate. \textbf{R}: canonicalization applied before indexing. \textbf{M}: successful-approach matching used in ranking. All rows share the frozen backbone, action proposer, controller, and 40-step budget.}
\label{tab:ablations}
\centering
\footnotesize
\setlength{\tabcolsep}{4.5pt}
\begin{tabular}{@{}cccc rr rr@{}}
\toprule
\multicolumn{4}{c}{Component} & \multicolumn{2}{c}{HM3D v0.1} & \multicolumn{2}{c}{HM3D v0.2}\\
\cmidrule(lr){1-4}\cmidrule(lr){5-6}\cmidrule(lr){7-8}
G & P & R & M & SR & SPL & SR & SPL\\
\midrule
$\times$ & $\times$ & $\times$ & $\times$ & 56.3 & 30.3 & 67.2 & 30.7\\
\checkmark & $\times$ & \checkmark & \checkmark & 58.3 & 31.2 & 70.9 & 31.1\\
\checkmark & \checkmark & $\times$ & \checkmark & 59.0 & 30.7 & 70.8 & 31.3\\
\checkmark & \checkmark & \checkmark & $\times$ & 59.1 & 29.0 & \textbf{71.0} & 30.2\\
\checkmark & \checkmark & \checkmark & \checkmark & \textbf{59.9} & \textbf{31.8} & \textbf{71.0} & \textbf{31.5}\\
\bottomrule
\end{tabular}
\end{table}

\begin{figure*}[t]
\centering
\includegraphics[width=1.0\textwidth]{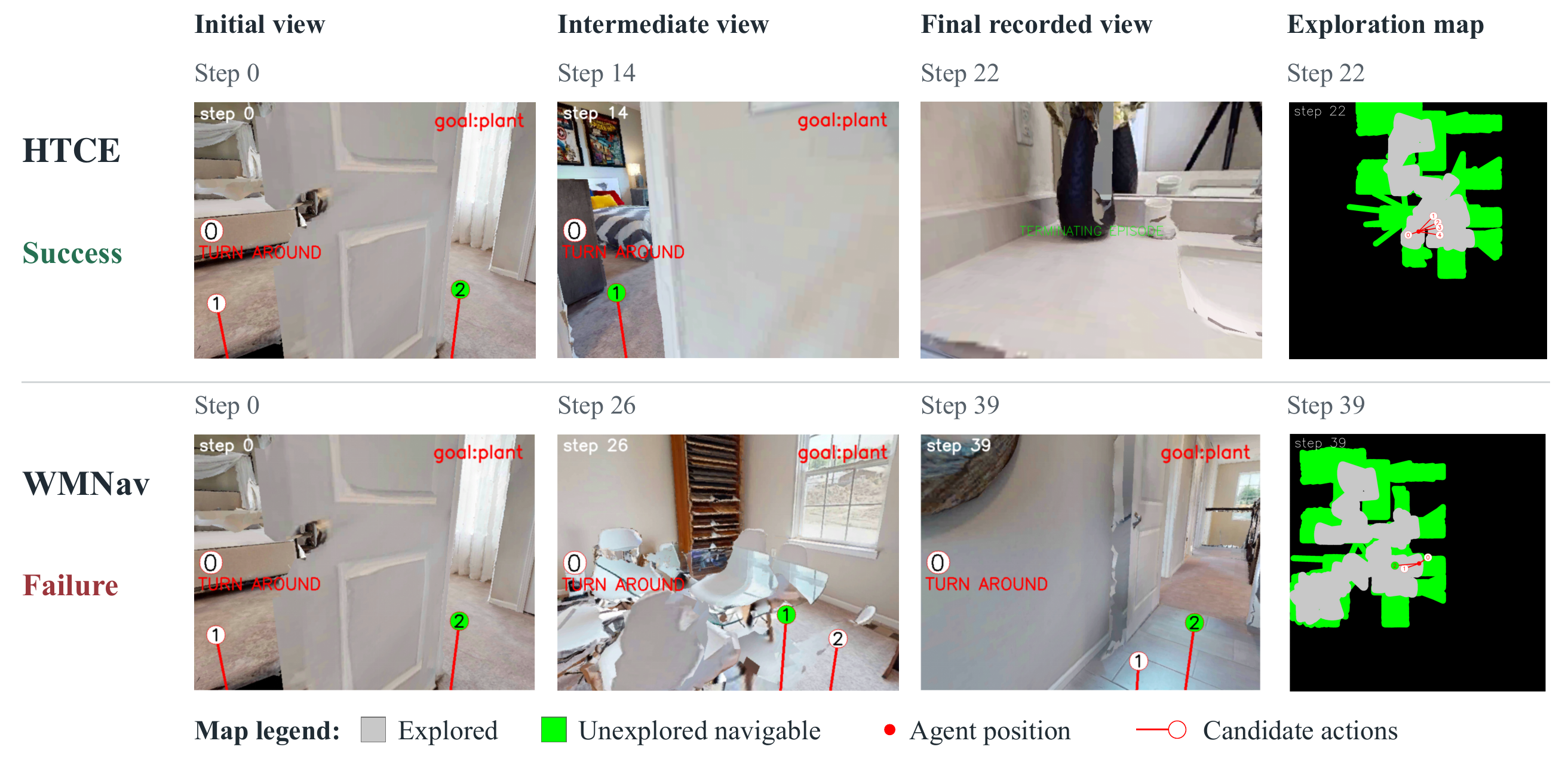}
\caption{Plant-search episodes in HM3D v0.2. Top: \method{} reaches the target at step~22. Bottom: WMNav fails to reach the target by step~39. Egocentric views show the initial, intermediate, and final recorded observations. In the exploration maps, gray denotes explored regions and green denotes mapped navigable regions not yet marked as explored. Red dots indicate agent positions; numbered rays are logged candidate actions, not executed trajectories.}
\label{fig:case}
\end{figure*}

\subsection{Route Granularity and Retained Semantics}
\label{sec:representation}
Transition relabeling, serial subdivision, and adjacent-anchor repetitions form 656 equivalent pairs. Destination changes, nonpalindromic reversal, and transition removal form 669 pairs with different patterns. All 1,325 pairs satisfy their prescribed exact-key relations, testing invariance to description and sampling changes alongside sensitivity to destination, direction, and passage relations. Appendix~\ref{app:diagnostics} specifies the deterministic construction.

Reduction turns partial agreement into exact agreement: raw edit similarities of 0.798, 0.721, and 0.796 for the three equivalent transformations all become 1.000, so serial passage observations no longer split a match. Destination, order, and connector changes instead yield distinct patterns, keeping the distinctions a route must preserve (Appendix~\ref{app:threshold}).

\subsection{Structural Properties of the Canonical Form}
\label{sec:graph_validation}
The reduction is also checked against controlled graph transformations that test invariance to compatible labels and subdivisions together with sensitivity to protected structure. Nine transformations over 36 branched contexts give 324 pairs, and an independent multigraph-isomorphism checker against Definition~\ref{def:sgh} satisfies 324/324 prescribed relations: relabeling a hallway step, inserting extra hallway steps, and the two combined leave the correspondence intact, while adding a branch, replacing an endpoint room, moving a branch leaf, or inserting a second target-room visit all remain distinct. Neither operation alone covers every case, so both are needed. The correspondence also holds over every connected graph shape on two to five vertices under 322 edge refinements (Appendix~\ref{app:coverage}). The continuous retrieval score grades the corresponding pairs as well: relabeling scores 1.000 and subdivision-related contexts 0.800--0.867, the gap carried entirely by the route-edge component recording suppressed-transition counts, so observation granularity remains a ranking attribute on top of the exact relation (Appendix~\ref{app:threshold}). The per-edit breakdown is given in Appendix~\ref{app:graph_tests} (Table~\ref{tab:graph_correspondence}).

\noindent\textit{Distinguishing branch position and multiplicity.}
Table~\ref{tab:hard_negatives} compares two structural changes. First, exchanging a kitchen branch at $j_1$ with a bathroom branch at $j_2$ on the ordered route $a$--$j_1$--$j_2$--$b$ changes which junction each branch is attached to. Second, changing two same-role leaves at one junction to three changes connection multiplicity. Each test pair consists of an original graph and its structurally non-equivalent edited graph. Three start roles, two destination roles, and two branch-role choices give $3\times2\times2=12$ pairs for each construction and 24 pairs in total; these are controlled graph pairs, not navigation episodes. Exact correspondence distinguishes all 24 pairs by retaining edge incidence, whereas the online descriptor $\Gamma$ is identical within each pair. The checkmarks and crosses summarize these outcomes across the tested label combinations. These examples identify branch position and multiplicity distinctions retained by Definition~\ref{def:sgh} but lost in its online projection; Appendix~\ref{app:hardneg} gives the construction details.

\begin{table}[t]
\caption{Structural distinctions preserved by exact correspondence but merged by the online descriptor.}
\label{tab:hard_negatives}
\centering\footnotesize
\setlength{\tabcolsep}{4pt}
\begin{tabularx}{\columnwidth}{@{}Y c c@{}}
\toprule
Structural change & \multicolumn{2}{c}{Distinguishes the pair?}\\
\cmidrule(lr){2-3}
 & Online $\Gamma$ & Exact $\sim_{\mathrm{SC}}$\\
\midrule
Branch positions swapped & $\times$ & $\checkmark$\\
\addlinespace[2pt]
Same-role leaves: $2\to3$ & $\times$ & $\checkmark$\\
\bottomrule
\end{tabularx}
\par\smallskip
\parbox{\columnwidth}{\footnotesize\raggedright
$\checkmark$: all tested pairs distinguished; $\times$: none.
Each pair contains an original graph and its structurally non-equivalent edited graph. Construction details are in Appendix~\ref{app:hardneg}.}
\end{table}

\subsection{Observation Provenance}
\label{sec:provenance}
Separating local from through-opening evidence lets a retrieved cue specify not only the observed target but the spatial relation it was seen from: local evidence binds a sighting to the observer region, while through-opening evidence records a view toward another region (Fig.~\ref{fig:method}(a)). Keeping the two distinct adds 0.6 and 0.4 SPL points over merging them (Table~\ref{tab:ablations}).

\subsection{Qualitative Navigation}
\label{sec:qualitative}
Fig.~\ref{fig:case} compares HM3D plant searches from the same initial doorway view: \method{} reaches the target at step 22, where the baseline is still exploring at step 39. The egocentric views and top-down maps show retrieved experience committing the search to a productive region early instead of expanding coverage.

\subsection{Computational Cost}
\label{sec:cost}

Table~\ref{tab:cost} reports the CPU-side retrieval latency, memory-update time, and largest serialized worker snapshot over two episode orders. Measurements are taken with prompt injection disabled and exclude VLM inference and network communication. Appendix~\ref{app:evaluation} specifies the measurement protocol.

\begin{table}[t]
\caption{Memory-component computational cost over two episode orders. Orders A and B use seeds 42 and 43, respectively. Prompt injection is disabled; VLM inference and network communication are excluded.}
\label{tab:cost}
\centering
\small
\begin{tabularx}{\columnwidth}{@{}Yrr@{}}
\toprule
Measurement & Order A & Order B\\
\midrule
Retrieval (ms/call) &0.139&0.132\\
Update (ms/episode) &61.6&62.6\\
Largest snapshot (bytes) &550,077&514,524\\
\bottomrule
\end{tabularx}
\end{table}

\section{Further Remarks}
\label{sec:discussion}
\subsection{Structure-Preserving Design}
The subdivision viewpoint identifies which graph changes should preserve an approach's structural identity. Semantic constraints make the operation useful for navigation: anchors retain region categories, junctions retain alternative connections, and query endpoints establish approach direction. Suppressing intermediate degree-two transitions exposes this retained structure, while recorded transition counts make observation granularity available as a graded attribute.

\subsection{Experience Reuse With a Frozen Backbone}
Experience accumulation extends a frozen VLM through explicit memory updates. Fixed semantic rules and graph traversal organize the records; edit distance and set similarity retrieve related approaches. Perception and action prompts share the same model, making the operations inspectable with measurable computational cost.

\subsection{Complementary Evaluation Evidence}
Region predictions and the transition vocabulary determine the comparison resolution. The navigation comparison, route consolidation, and controlled graph transformations address complementary aspects of experience reuse: task performance, compact description, and structural correspondence.

\subsection{Scope of the Construction}
Two parameters set where the framework applies. The semantic role map fixes the comparison resolution: labels placed in the transition class are the ones a reduction is licensed to absorb, so the partition is the interface through which a deployment states which distinctions matter to it. In indoor navigation we instantiate it with passage labels and non-committal region labels, both of which carry no navigational commitment and are therefore the right candidates for suppression; a deployment that needs those labels preserved moves them into the anchor class and obtains a correspondingly finer correspondence, with Definition~\ref{def:sgh} and Proposition~\ref{prop:topology} holding unchanged for any such partition. The descriptor resolution is the second parameter, graded between the fast projection used online and the exact relation that defines the structural ground truth. Both are explicit choices with measured consequences, which is what allows the framework to be matched to applications beyond the residential-navigation setting evaluated here.

\section{Conclusion}
\label{sec:conclusion}
\method{} makes navigation experience comparable across scenes through homeomorphism-inspired topological canonicalization. Its structural correspondence absorbs compatible feature and topology refinements while retaining semantic anchors, chain words, branch incidence, and ordered query endpoints. The resulting reduction is a lossless compression of navigation experience: it merges 17.7--23.6\% of distinct stored route descriptions, satisfies every prescribed invariance and sensitivity relation across the enumerated transformations, and retains full task performance. Built on this index, \method{} attains the best SR and SPL in all three navigation settings under a shared frozen backbone, and structured experience retrieval accounts for up to 2.8 SPL points of the margin. Canonical structure, rather than scene-specific geometry, is what makes navigation experience transferable.

\label{sec:main_end}

\FloatBarrier
\bibliographystyle{IEEEtran}
\bibliography{references}
\clearpage
\appendices
\section{Implementation Details}
\label{app:implementation}
\label{app:coefficients}
Three constraints set the coefficients below rather than a per-benchmark search: the weights of each composite score are nonnegative and sum to one, so all scores share the range $[0,1]$; a term that should outrank another carries the larger weight; and indirect or prior-like evidence is bounded so that it settles near-ties without overturning direct evidence.

\subsection{Observation Association}
Each observation records the predicted region, visible objects, visual summary, pose estimate, previous action, target, and episode/step identifiers. Local and through-opening objects are stored separately. Region association uses
\begin{align}
q_{\mathrm{sem}}&=0.45q_{\mathrm{name}}+
0.35\min(1,J_{\mathrm{in}}+0.3J_{\mathrm{rem}})
\nonumber\\
&\quad+0.20J_{\mathrm{desc}},\\
q_{\mathrm{assoc}}&=\operatorname{clip}
(q_{\mathrm{sem}}+q_{\mathrm{topo}}+q_{\mathrm{sp}},0,1).
\end{align}
Name agreement credits an exact name above a stored alias and nothing for any other label, and the $J$ terms compare object sets and description keywords, discounting through-opening objects because they may belong to a neighboring region. The three semantic weights decrease with how specifically each term identifies a region: a region name names it directly, object overlap is shared by rooms of the same type, and free-text keywords are the least constrained. The topology term rewards the previous region and its neighbors and penalizes a non-neighbor whose semantic agreement is low, while the spatial term decays with $d_{\min}$, the distance in meters from the current position estimate to the nearest recent position stored for the candidate region. These two corrections contribute at most $0.12+0.08=0.20$, the weight of the weakest semantic term, so they settle near-ties but cannot override semantic disagreement. A score above the confirmation threshold accepts the association, an intermediate score or a narrow same-name margin gives a tentative assignment, and a new region requires either sufficient confidence or two repeated unmatched observations. Table~\ref{tab:configuration} lists these values.

\begin{table}[t]
\caption{Fixed implementation configuration. The same values are used for every dataset, split, and ablation. Scoring weights appear in (\ref{eq:graph_weights_impl})--(\ref{eq:structural_support_impl}); Appendix~\ref{app:coefficients} states the constraints that set them.}
\label{tab:configuration}
\centering\footnotesize
\setlength{\tabcolsep}{4pt}
\begin{tabularx}{\columnwidth}{@{}Yl@{}}
\toprule
Setting & Value\\
\midrule
\multicolumn{2}{@{}l}{\emph{Region association}}\\
Name agreement: exact / alias / other & 1 / 0.7 / 0\\
Topology term: previous / neighbor / non-neighbor & 0.05 / 0.12 / $-0.08$\\
Semantic agreement below which the non-neighbor term applies & 0.70\\
Recent-position term & $\max(0,0.08(1-d_{\min}/5))$\\
Confirmed association & $\geq0.60$\\
Tentative association & $[0.45,0.60)$\\
Tentative same-name margin & $<0.08$\\
New-region confidence & 0.70\\
\addlinespace
\multicolumn{2}{@{}l}{\emph{Stored records}}\\
Region observations / object viewpoints per history & 20 / 10\\
Region identifiers per stored approach & 12\\
Branch-tracing budget in transition hops & 8\\
\addlinespace
\multicolumn{2}{@{}l}{\emph{Retrieval ranking}}\\
Structural-support weight $\lambda$ & 0.04\\
Gate: association / path confidence & 0.75 / 0.65\\
Gate: reduced-route length $\ell$ & $\leq4$\\
Same-region support & $h=0.85$\\
Missing path or failed gate & $h=0$\\
\bottomrule
\end{tabularx}
\end{table}

\subsection{Object Evidence and Success Records}
Object records retain confidence, observation count, first/last episode, viewpoints, and successful approaches. Viewpoints contain episode/step identifiers, pose, visual summary, and navigation hint. Deduplication uses episode and step, and each history keeps a bounded number of entries (Table~\ref{tab:configuration}). Through-opening records additionally retain the observer region, relation type, and likely target region; object-conditioned phrases in the VLM summary identify the relation, and a compatible neighbor supplies the target-region link when available. The policy receives selected evidence as text alongside its current image.

Terminal success attaches the approach to the reached local target record, storing a bounded number of its most recent region identifiers, the full reduced role sequence, and the available graph signature. Successful signatures use the graph at write time; query paths use the current graph.

\subsection{Retrieval Scores and Parameters}
\label{app:retrieval_config}
The six components $\phi_i$ in (\ref{eq:graph_score}) and their fixed weights $w_i$ are:
\begin{equation}
\begin{aligned}
S_{\mathrm{graph}}={}&0.30\,S_R+0.20\,J_E+0.20\,J_B\\
&+0.15\,J_s+0.10\,J_g+0.05\,J_N,
\end{aligned}
\label{eq:graph_weights_impl}
\end{equation}
where $S_R$ is edit similarity on retained role sequences and each $J$ is Jaccard overlap of the corresponding descriptor set (route edges, branches, start-endpoint neighbors, target-endpoint neighbors, and node descriptors). The weights sum to one, so $S_{\mathrm{graph}}$ shares the range of the feature-word score of (\ref{eq:route_score}), and they decrease with each component's contribution to approach identity: the retained role sequence carries the ordered approach itself, route-edge and branch summaries describe the structure hanging off it, endpoint neighborhoods add local context, and node descriptors rank last because the preceding terms already express much of their role and degree information.
The base scores are
\begin{align}
b_{\mathrm{in}}(v,g)
&=c_o+0.03n_o+0.20n_o^+
+0.20\ind[v\text{ confirmed}],\\
b_{\mathrm{rem}}(v,g)
&=c_o+0.02n_o+0.12\ind[v\text{ confirmed}]
\nonumber\\
&\quad+0.08\ind[\text{linked target region}],
\end{align}
where $c_o$ is object confidence, $n_o$ observation count, and $n_o^+$ the number of successful records. These scores rank candidate destinations rather than estimate probabilities: object confidence supplies the base value and the remaining terms are bonuses encoding two orderings. A prior successful approach or a confirmed association each add $0.20$, more than six further sightings contribute, so demonstrated success outranks observation count. Through-opening evidence uses smaller bonuses than local evidence ($0.02$ against $0.03$ per sighting, $0.12$ against $0.20$ on confirmation) because its region attribution is indirect, while an explicit target-region link restores part of the gap. Local target records take priority, the implementation excludes an unconfirmed toilet record outside a bathroom unless it has successful support, and an empty candidate set returns no historical target cue.

Target-compatible region roles $\mathcal R_g$ are fixed: chairs favor dining rooms, offices, bedrooms, and living rooms; toilets favor bathrooms; sofas favor living rooms; beds favor bedrooms; plants favor living rooms, hallways, bedrooms, and offices; television screens favor living rooms, bedrooms, and media rooms.

For a candidate in another region, the support function in (\ref{eq:structural_term}) is
\begin{equation}
\begin{aligned}
\psi(v,g)=\operatorname{clip}\bigl[
&\bar q e^{-(\ell-1)/2}(0.5+0.5S)\\
&+0.08\ind[r(f(v))\in\mathcal R_g],0,1\bigr].
\end{aligned}
\label{eq:structural_support_impl}
\end{equation}
Here $\bar q$ denotes mean path confidence, $\ell=|s(\pi)|$ the full reduced-route length, and $S$ graph or feature agreement. Ranking uses the structural-support weight $\lambda$ of Table~\ref{tab:configuration}. Gate $G$ requires association and path confidence above their thresholds, a reduced-route length within the retrieval horizon, and successful support. Same-region candidates receive a fixed support value after association and success checks; missing paths or failed gates give $h=0$. Matching uses graph signatures when enabled and available, otherwise successful role sequences. Exact score ties retain insertion order. Each factor carries one constraint: $0.5+0.5S$ maps structural agreement into $[0.5,1]$, so a weak match attenuates path support instead of eliminating it; the exponential multiplies support by $e^{-1/2}\approx0.61$ per additional reduced step, favoring short supported approaches; and role compatibility is a prior worth at most $0.08$ against a path term reaching $1$. Because $h\in[0,1]$, $\lambda$ bounds the total influence of structural support at $0.04$ of a candidate's base score, so retrieval reorders comparable candidates but never overrides object evidence. The structural properties of Section~\ref{sec:method} do not depend on these ranking coefficients, which belong to the retrieval configuration.

\subsection{Reduction, Signatures, and Complexity}
\label{app:reduction}
\label{app:signature_config}
Region normalization lowercases names, replaces underscores, collapses whitespace, and applies a fixed alias map including hall to hallway, lounge/sitting room to living room, bed room/sleeping room to bedroom, and toilet/restroom/washroom to bathroom. Empty or uncertain labels map to unknown room, and other labels are retained as semantic anchors.

Online matching uses full contracted role sequences, and runtime retrieval compares individual successful records. Pattern analysis instead bounds each route by its first and last anchors and serializes direct/transition connectors; anchorless sequences have no key, and Table~\ref{tab:boundary} gives the resulting relations. Suppression uses stored-graph degrees and protects query endpoints. Branch tracing follows a bounded number of transition hops (Table~\ref{tab:configuration}), stopping at a chain endpoint or blocked route node. Role sequences preserve order, while descriptor sets collapse repetitions, so degree buckets and sets can coincide for distinct graphs (Table~\ref{tab:hard_negatives}).

For a route of length $L$, sequence reduction costs $O(L)$. Edit comparison costs $O(K_1K_2)$ for reduced lengths $K_1,K_2$ \cite{wagner1974string}. Comparing $n_+$ successful sequences of maximum length $K$ costs $O(n_+K^2)$. BFS over an adjacency list costs $O(|V|+|E|)$; the implemented neighbor lookup scans the edge dictionary, giving an $O(|V||E|)$ worst-case query cost in addition to path copying. Bounded record lists limit the comparison count.

\section{Semantic and Topological Reduction Properties}
\label{app:proof}
\subsection{Canonical Feature Words}
Consider normalized, anchor-bounded words over disjoint anchor and transition alphabets. Replacing a transition name by $\tau$ reduces the number of noncanonical symbols. Either contraction in (\ref{eq:rewrite}) reduces length. The lexicographic measure given by these two quantities therefore decreases at each reduction.

Disjoint contractions commute. Overlaps occur within a run $\tau\tau\tau$ or $aaa$ of one anchor; either contraction yields the same two-symbol run and then the same singleton. Role replacement has a unique output and does not overlap anchor replacement because the alphabets are disjoint. Termination and local confluence give confluence and a unique normal form \cite{newman1942,baader1998term}.

Typed serialization records the ordered anchors and the presence or absence of an intervening $\tau$ and is injective on reduced anchor-bounded words, so different reduced anchor or connector sequences have different keys. A nonpalindromic reversal changes the pattern; a palindromic reversal leaves it unchanged. Table~\ref{tab:boundary} gives representative cases.

\begin{table}[t]
\caption{Route-pattern relations. Distinct letters denote distinct semantic anchors and $t_i$ transition observations.}
\label{tab:boundary}
\centering\footnotesize
\begin{tabularx}{\columnwidth}{@{}l l Y@{}}
\toprule
Inputs & Relation & Preserved information\\
\midrule
$a\,t_1\,b$, $a\,t_2\,b$ & Same & Transition-role normalization\\
$a\,t_1\,b$, $a\,t_1\,t_2\,b$ & Same & Passage subdivision\\
$a\,a\,t_1\,b$, $a\,t_1\,b$ & Same & Adjacent repetition\\
$a\,t_1\,b$, $a\,t_1\,c$ & Different & Destination category\\
$a\,t_1\,b$, $b\,t_1\,a$ & Different & Nonpalindromic order\\
$a\,t_1\,b$, $a\,b$ & Different & Connector type\\
$a\,t_1\,a$, $a$ & Different & Separated visits\\
$t_1\,a\,t_2$, $a$ & Same after cropping & Anchor-bounded domain\\
$t_1\,t_2$ & No key & Anchor required\\
\bottomrule
\end{tabularx}
\end{table}

\subsection{Topological Reduction and Preserved Incidence}
\label{app:topology_proof}
\emph{Proof of Proposition~\ref{prop:topology}.}
Every unprotected vertex is a transition with two incident edges. Starting from a protected vertex, following such vertices determines a unique maximal chain ending at a protected vertex, possibly the starting vertex itself. The rooted connected context contains protected endpoints; a transition-only cycle is therefore either split by these endpoints or attached through a protected junction. These chains partition the edges. Replacing each by an edge retains its two endpoint incidences; loops and parallel edges retain returning connections and alternative passages.

Every original walk between protected vertices projects to a walk in $\mathcal K$, and expanding reduced edges lifts it back. Reachability is consequently preserved in both directions. Every off-route branch remains incident on its original protected vertex, and route-edge marks retain traversal order. Suppression inside one chain changes neither another chain nor its endpoints; any suppression order yields the same chain partition. Compatible relabeling preserves roles, and admissible subdivision preserves each canonical chain word by Proposition~\ref{prop:route}, which establishes the claimed invariance. Identity, inverse, and composition of the preserving isomorphisms also make $\sim_{\mathrm{SC}}$ an equivalence relation on the specified contexts.

\subsection{Protected Structure and Counterexamples}
\label{app:protected_cases}
Semantic anchors preserve region identity, junctions preserve branch incidence, and query endpoints preserve the directed task. Feature contraction acts only on descriptions: two distinct adjacent bedrooms $b_1,b_2$ both survive even though the feature word contracts their repeated category, because identifying them in the graph would also combine their separate doors and evidence. Likewise, removing the transition in $a$--$\tau$--$b$ preserves bare connectivity but changes its chain word to $a$--$b$, so a transition endpoint is protected even at degree two. At a cropped context boundary, the stored degree and cut mark preserve the distinction between an observed branch and an interior passage.

The branch-attachment and multiplicity examples in Section~\ref{sec:graph_validation} give explicit pairs with identical $\Gamma$ and different $\mathcal K$ (Table~\ref{tab:hard_negatives}), so descriptor equality and structural correspondence have different meanings: the exact relation retains the full incidences, while the retrieval descriptor summarizes them through degree buckets and sets.

\section{Structural Analysis Details}
\label{app:diagnostics}
\subsection{Controlled Graph Constructions}
\label{app:graph_tests}
The branched contexts in Section~\ref{sec:graph_validation} combine living-room, dining-room, or office starts; bedroom or bathroom targets; kitchen or study branches; and $k\in\{1,2,3\}$ hallway transitions, where the start, target, and branch leaf are semantic anchors. Positive edits replace hallway with corridor, insert two hallways on $(s,p_1)$, or combine both. Negative edits add a kitchen at $p_1$, change the start to garage or target to media room, move the endpoint to $b$, reattach $b$ from $j$ to $s$, or insert a distinct target-role vertex before $t$.

The NetworkX 3.2.1 multigraph checker preserves roles, boundaries, route order, chain words, and edge multiplicity. In the graph-shape sweep, the first and last indexed vertices are living room and bedroom with a shortest query path, and other vertices are either all transitions or office at odd indices and transition at even indices; each edge receives two hallway subdivisions, and chain-word equality determines the expected relation.

Each row of Table~\ref{tab:graph_correspondence} applies one edit to all 36 contexts and asks whether the edited graph still describes the same approach, with the answer fixed before matching. Four reference reductions isolate the two operations: raw labels and vertices (R), role normalization only (F), suppression of intermediate hallway steps only (T), and both (S). A reduction is correct when it returns the prescribed answer, and every row is uniform across the 36 pairs, so correct entries are marked \checkmark{} and incorrect ones $\times$. The last column averages $S_{\mathrm{graph}}$.

\begin{table}[t]
\caption{Structural correspondence under controlled graph edits, 36 contexts per row. \checkmark{} correct on all 36, $\times$ correct on none. R: raw labels; F: role normalization only; T: suppression of intermediate hallway steps only; S: both (structural correspondence).}
\label{tab:graph_correspondence}
\centering\footnotesize
\setlength{\tabcolsep}{4pt}
\begin{tabularx}{\columnwidth}{@{}Y c cccc r@{}}
\toprule
Graph edit & Same & R & F & T & S & Mean\\
 & approach? & & & & & score\\
\midrule
Hallway step relabeled corridor & Yes & $\times$ & \checkmark & $\times$ & \checkmark & 1.000\\
Extra hallway steps inserted & Yes & $\times$ & $\times$ & \checkmark & \checkmark & 0.867\\
Both of the above & Yes & $\times$ & $\times$ & $\times$ & \checkmark & 0.867\\
\addlinespace
Kitchen branch added at a hallway step & No & \checkmark & \checkmark & \checkmark & \checkmark & 0.725\\
Start room replaced & No & \checkmark & \checkmark & \checkmark & \checkmark & 0.742\\
Target room replaced & No & \checkmark & \checkmark & \checkmark & \checkmark & 0.742\\
Route ends at the branch leaf & No & \checkmark & \checkmark & \checkmark & \checkmark & 0.542\\
Branch leaf moved to another node & No & \checkmark & \checkmark & \checkmark & \checkmark & 0.213\\
Second target-room visit inserted & No & \checkmark & \checkmark & \checkmark & \checkmark & 0.746\\
\bottomrule
\end{tabularx}
\end{table}

\subsection{Recorded Routes}
\label{app:recorded_routes}
Orders A/B use seeds 42/43. Records contain worker/episode identifiers, target, raw and final region labels, and a stored skeleton. The analysis uses the 483/444 successful-route records that contain at least one semantic anchor; empty and transition-only routes have no anchor-bounded key.

All reductions use final labels, which incorporate region association, yielding 130/140 anchor-bounded sequences and 107/107 patterns. Transition-role mapping accounts for the first part of the merge (130/140 to 123/132) and consecutive-role contraction for the remainder, giving index reductions of 17.7\% and 23.6\%. Including transition endpoints gives 145/147 role sequences from 217/230 final-label sequences. This common input isolates the representation operation, and all pattern counts and similarity scores use the same visible final region-label sequences.

\subsection{Deterministic Transformations}
The sampling units are the distinct anchor-bounded inputs within each order. For each eligible sequence we apply the six transformations named in Table~\ref{tab:perturbations}: \emph{transition alias} replaces the first transition with another label in the fixed transition vocabulary; \emph{transition subdivision} duplicates the first transition observation twice; \emph{adjacent anchor repeat} repeats the first anchor next to itself; \emph{destination change} replaces the last anchor with bedroom, or with kitchen when it is already bedroom; \emph{order reversal} reverses the sequence and retains the pair when its reduced key changes; and \emph{transition removal} deletes the first maximal transition run. The first three preserve the declared route pattern and the last three change it. Base routes can recur across conditions and orders; these are deterministic mechanism checks with descriptive counts.

\begin{table}[t]
\caption{Exact route-pattern decisions on all 1,325 deterministic transformations.}
\label{tab:perturbations}
\centering\footnotesize
\begin{tabularx}{\columnwidth}{@{}Ylrr@{}}
\toprule
Transformation & Pattern & Pairs & Correct\\
\midrule
Transition alias & Same &193&193\\
Transition subdivision & Same &193&193\\
Adjacent anchor repeat & Same &270&270\\
Destination change & Different &270&270\\
Order reversal & Different &206&206\\
Transition removal & Different &193&193\\
\bottomrule
\end{tabularx}
\end{table}

\begin{table}[t]
\caption{Accepted pairs at edit similarity $\geq0.6$. The first three conditions are equivalent routes; the last three have different patterns. This is an offline diagnostic.}
\label{tab:threshold_diagnostics}
\centering\footnotesize
\begin{tabularx}{\columnwidth}{@{}Yrrr@{}}
\toprule
Transformation & Pairs & Raw & Reduced\\
\midrule
Transition alias &193&193&193\\
Transition subdivision &193&193&193\\
Adjacent anchor repeat &270&257&270\\
Destination change &270&230&228\\
Order reversal &206&45&50\\
Transition removal &193&152&156\\
\bottomrule
\end{tabularx}
\end{table}

\subsection{Hard-Negative Structural Pairs}
\label{app:hardneg}
The two constructions of Section~\ref{sec:graph_validation} are instantiated over three start roles ($a\in\{$living room, dining room, office$\}$), two destination roles ($b\in\{$bedroom, bathroom$\}$), and two branch-role choices, each combination giving one pair of a base graph and its edited partner. The 12 instances per construction vary only these labels, so the reported decisions cover the tested combinations rather than arbitrary ones, and the counts are multiples of 12 by construction. The $\checkmark$ and $\times$ entries of Table~\ref{tab:hard_negatives} report descriptor equality versus exact structural correspondence, not thresholded retrieval decisions. A separate positive control inserts one transition vertex on the first route edge while leaving all other evidence unchanged, so it preserves structural correspondence and changes only the suppressed-transition count; these controls are excluded from the 24 pairs, and the table reports negative-pair outcomes only.

\subsection{Coverage Across Graph Shapes}
\label{app:coverage}
We further examine every connected simple graph shape with two to five vertices. The 30 shapes receive two fixed role assignments, yielding 60 rooted contexts, and inserting two transition vertices on each edge in turn gives 322 edge-refinement checks. All satisfy the expected semantic relation: extending an existing passage preserves correspondence, while introducing the first transition between directly adjacent anchors changes the chain word.

Named graph cases test the additional structures in Table~\ref{tab:graph_coverage}, where parallel passages and returning cycles retain their distinct edge incidences and cropped junctions retain their stored degree and boundary mark. Separate tests preserve results under vertex renaming and insertion-order changes. Together, these checks exercise the reachability and incidence properties in Proposition~\ref{prop:topology} over the stated sizes, role assignments, and edits.

\begin{table}[t]
\caption{Structural coverage of the independent structural-correspondence checker. The graph-shape sweep comprises 322 edge refinements; named cases test the additional boundary conditions below.}
\label{tab:graph_coverage}
\centering\footnotesize
\begin{tabularx}{\columnwidth}{@{}lY@{}}
\toprule
Graph context & Verified property\\
\midrule
Small-graph edge refinements & Correspondence follows semantic chain-word preservation\\
Parallel passages & Distinct connections retain multiplicity\\
Returning cycles & Loops retain their protected attachment\\
Transition endpoints & Start and target remain explicit\\
Cropped junctions & Stored degree and boundary mark remain explicit\\
Distinct same-role anchors & Separate region vertices remain distinct\\
Single-vertex query & The rooted context is retained\\
\bottomrule
\end{tabularx}
\end{table}

\subsection{Continuous Similarity}
\label{app:threshold}
Table~\ref{tab:threshold_diagnostics} gives complete acceptance counts at the offline threshold 0.6 for raw label sequences and contracted role sequences, separating pattern identity from continuous similarity. At the same threshold, 144 of 216 structurally non-equivalent graph pairs satisfy $S_{\mathrm{graph}}\geq0.6$, while the exact checker distinguishes all 216. Online retrieval uses graded scores and evidence gates.

\section{Evaluation Protocol}
\label{app:evaluation}
\subsection{Navigation Settings}
The navigation experiments use Qwen3-VL-Plus with a 40-step interaction budget. The prompt-context and direct-action-bias interfaces are enabled in the main \method{} configuration, with retrieved guidance applied subject to evidence confidence and action-selection conditions. The retrieval configuration of Table~\ref{tab:configuration} is shared by all three datasets and by every component experiment; no coefficient, threshold, or gate is adjusted per dataset or per split. The component experiments vary successful-approach support, route reduction, and observation provenance as described in Section~\ref{sec:components}.

\subsection{Episode Scheduling and Memory}
Navigation episodes are assigned to parallel workers by dataset ordering and evaluation seed; for the HM3D v0.2 route analyses, two episode orders generated with seeds 42 and 43 each assign 1,000 episodes to 50 workers. Persistent memory stores semantic region observations, graph connectivity, target evidence, and successful approaches across episode boundaries, with workers loading the shared snapshot at episode start and writing their updated snapshot at termination.

\subsection{Computational Cost}
Retrieval and memory-update costs are measured over two evaluation orders containing 985 and 964 episode events. Retrieval latency is total retrieval time divided by the number of calls, update time is averaged per episode, and memory footprint is the largest serialized worker snapshot. Measurements characterize the memory component with prompt injection disabled and exclude VLM inference and network communication.

\end{document}